\documentclass[letterpaper]{article} 
\usepackage{aaai25}  
\usepackage{times}  
\usepackage{helvet}  
\usepackage{courier}  
\usepackage[hyphens]{url}  
\usepackage{graphicx} 
\usepackage{natbib}  
\usepackage{caption} 
\usepackage{multirow}

\usepackage{amsmath,amssymb}
\usepackage{booktabs} 
\usepackage{float} 
\usepackage{lipsum}   

\usepackage{algorithm}
\usepackage{algorithmic}
\usepackage{xcolor}
\usepackage{footmisc}

\usepackage{newfloat}
\usepackage{listings}
\DeclareCaptionStyle{ruled}{labelfont=normalfont,labelsep=colon,strut=off} 
\floatstyle{ruled}
\newfloat{listing}{tb}{lst}{}
\floatname{listing}{Listing}
\title{Uncertainty-aware Knowledge Tracing}
\author{
    Weihua Cheng\textsuperscript{\rm 1},
    Hanwen Du\textsuperscript{\rm 2},
    Chunxiao Li\textsuperscript{\rm 3},
    Ersheng Ni\textsuperscript{\rm 4},
    Liangdi Tan\textsuperscript{\rm 1},
    Tianqi Xu\textsuperscript{\rm 3},
    Yongxin Ni\textsuperscript{\rm 3}\thanks{Corresponding author.}
}
\affiliations{
    \textsuperscript{\rm 1}ShanghaiTech University, Shanghai, China\\
    \textsuperscript{\rm 2}Soochow University, Suzhou, China\\
    \textsuperscript{\rm 3}University of Science and Technology of China, Hefei, China\\
    \textsuperscript{\rm 4}The University of Queensland, Brisbane, Australia\\
    chengwh2024@shanghaitech.edu.cn, hwdu@stu.suda.edu.cn,
    chunxiao.li@ustc.edu.cn,
    nelson.2044474491@outlook.com,
    tanld2022@shanghaitech.edu.cn,
    xutianqi@mail.ustc.edu.cn,
    niyongxin2016@gmail.com 
}

\usepackage{bibentry}

\begin{document}

\maketitle

\begin{abstract}
Knowledge Tracing (KT) is crucial in education assessment, which focuses on depicting students' learning states and assessing students' mastery of subjects. With the rise of modern online learning platforms, particularly massive open online courses (MOOCs), an abundance of interaction data has greatly advanced the development of the KT technology. Previous research commonly adopts deterministic representation to capture students' knowledge states, which neglects the uncertainty during student interactions and thus fails to model the true knowledge state in learning process. In light of this, we propose an Uncertainty-Aware Knowledge Tracing model (UKT) which employs stochastic distribution embeddings to represent the uncertainty in student interactions, with a Wasserstein self-attention mechanism designed to capture the transition of state distribution in student learning behaviors. Additionally, we introduce the aleatory uncertainty-aware contrastive learning loss, which strengthens the model's robustness towards different types of uncertainties. Extensive experiments on six real-world datasets demonstrate that UKT not only significantly surpasses existing deep learning-based models in KT prediction, but also shows unique advantages in handling the uncertainty of student interactions.

\end{abstract}
\begin{links}
\link{Code}{https://github.com/UncertaintyForKnowledgeTracing/UKT}
\end{links}
\section{Introduction}
\textit{Knowledge Tracing} (KT) is crucial in educational assessment, which tracks students' knowledge mastery by modeling dynamic knowledge states through continuous analysis of student interactions. It aids educators and platforms in evaluating abilities and personalizing learning paths for deeper knowledge absorption. With the growth of online education, platforms like MOOCs have generated vast interaction data, enriching training material for machine learning models. Nevertheless, the learning process is prolonged and variable, with each interaction characterized by uncertainty. This underscores the need for constructing KT models that can effectively handle the complexity of interactions.
\begin{figure}
		\centering 
            \includegraphics[width=\linewidth]{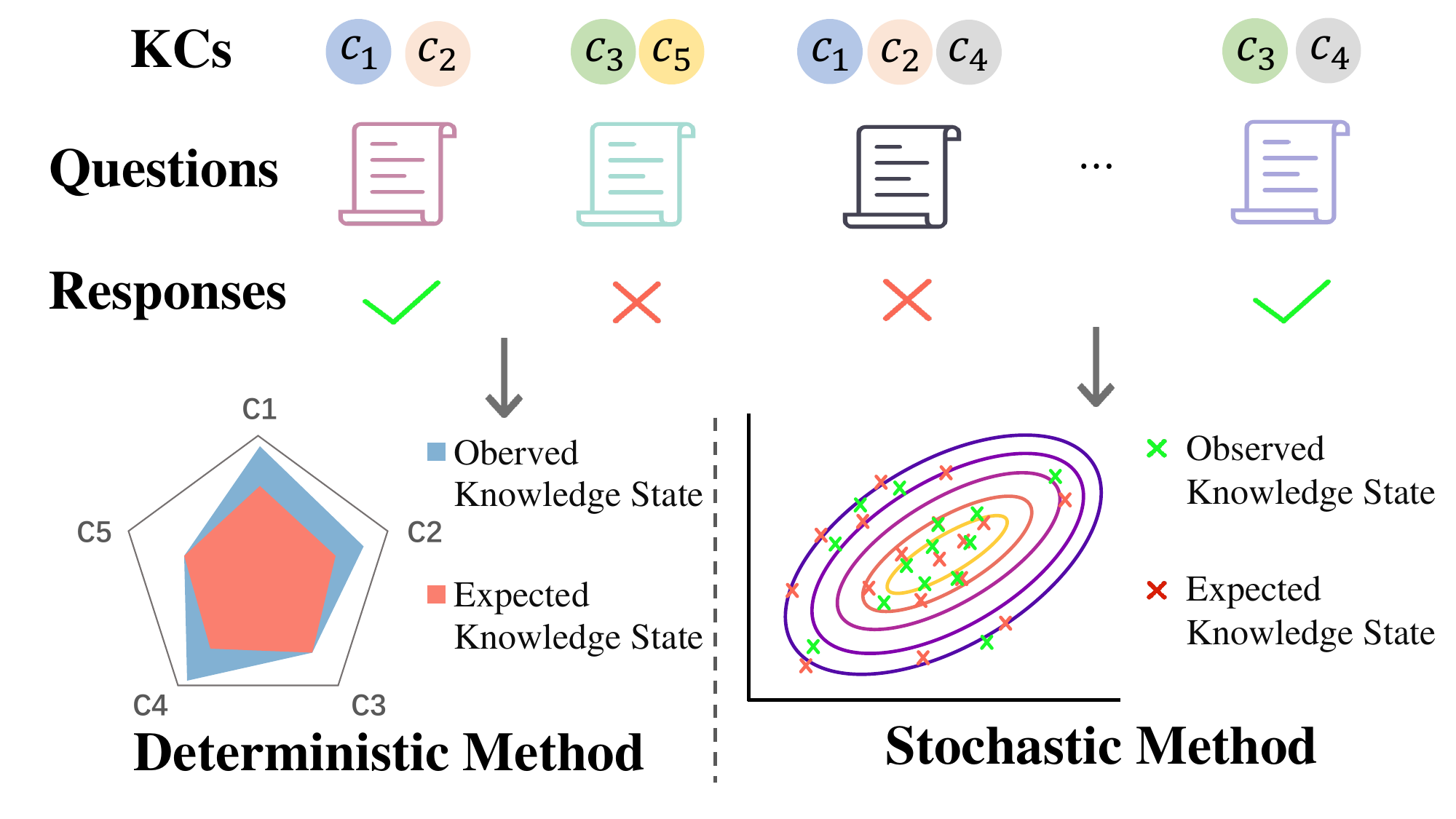}
		\caption{The impact of uncertainty on knowledge state under deterministic and stochastic modeling.} \label{differemt}
\end{figure}
Previous studies often use fixed embeddings to model student-course interactions. For instance, DKT \cite{piech2015deep} converts student interactions into fixed-length vectors using one-hot encodings or compressed representations. AKT \cite{choi2020towards} integrates concept embeddings, question difficulty, and response embeddings to capture the relationship between students' knowledge and questions. Additionally, various methods have emerged to apply neural components to capture forgetting behaviors \cite{nagatani2019augmenting}, recency effects \cite{zhang2021multi}, and other auxiliary information, e.g., question-KC relations \cite{pandey2020rkt, yang2021gikt}, question text \cite{liu2019ekt}, and learning ability \cite{shen2022assessing}. However, these methods assume that student behavior is solely determined by their current knowledge level, as they simply infer students' knowledge state from previous answers but overlook the stochastic factors in the learning process.

In educational environments, various factors can affect student performance, leading to uncertainties in knowledge assessments. These uncertainties arise from differences in learning ability, adaptability, and personal circumstances. Among them, \textbf{\textit{epistemic uncertainty}} helps models more accurately assess a student's true knowledge level, reflecting cognitive differences. In contrast, factors such as careless errors or lucky guesses are sourced from \textbf{\textit{aleatory uncertainty}}, which may mislead model assessment.
Such uncertainties are commonplace, given that students often demonstrate diverse levels of knowledge within the same sequence of interactions, with instances of misconduct that frequently deviate from their true abilities. As shown in Figure \ref{differemt}, differences in learning ability can cause deviations from the expected knowledge state in deterministic methods.

Capturing uncertainties can enable more personalized assessments of student learning. However, not all uncertainties are beneficial for learning assessments. 
While capturing epistemic uncertainty aids in improving assessments, the introduction of other uncertainties, i.e., aleatory uncertainty, could potentially lead models to make biased judgments in the opposite direction. For instance, a student may guess the correct answer on a multiple-choice question or make calculation errors on problems they could otherwise solve correctly due to carelessness. Such aleatory uncertainty does not accurately reflect students' true knowledge levels, and including these interactions could lead models to incorrect conclusions. Therefore, to ensure an accurate representation of students' knowledge mastery, it is crucial to enhance the model's robustness against aleatory uncertainty while retaining the epistemic uncertainty.

To tackle these problems, two key questions should be considered: 
\begin{itemize}
    \item How to model students' knowledge mastery with their uncertainties taken into consideration?
    \item How to strengthen model robustness towards aleatory uncertainty while retaining the epistemic one?
\end{itemize}

In light of this, this paper introduces Uncertainty-aware Knowledge Tracing (UKT) model to capture effective uncertainties in student behaviors. Specifically, 
UKT employs a stochastic method to depict the interaction history of each student as a Gaussian distribution, where the fundamental knowledge level and uncertainties of each student are quantified by the mean and covariance respectively. Additionally, to capture adjacent state transitions that represent the learning process, UKT proposes a Wasserstein distance \cite{ruschendorf1985wasserstein} based self-attention mechanism tailored for knowledge tracing, modeling relationships between distributions to monitor global changes rather than being limited to comparisons of individual points. 

\begin{figure}
		\centering
            \includegraphics[width=\linewidth]{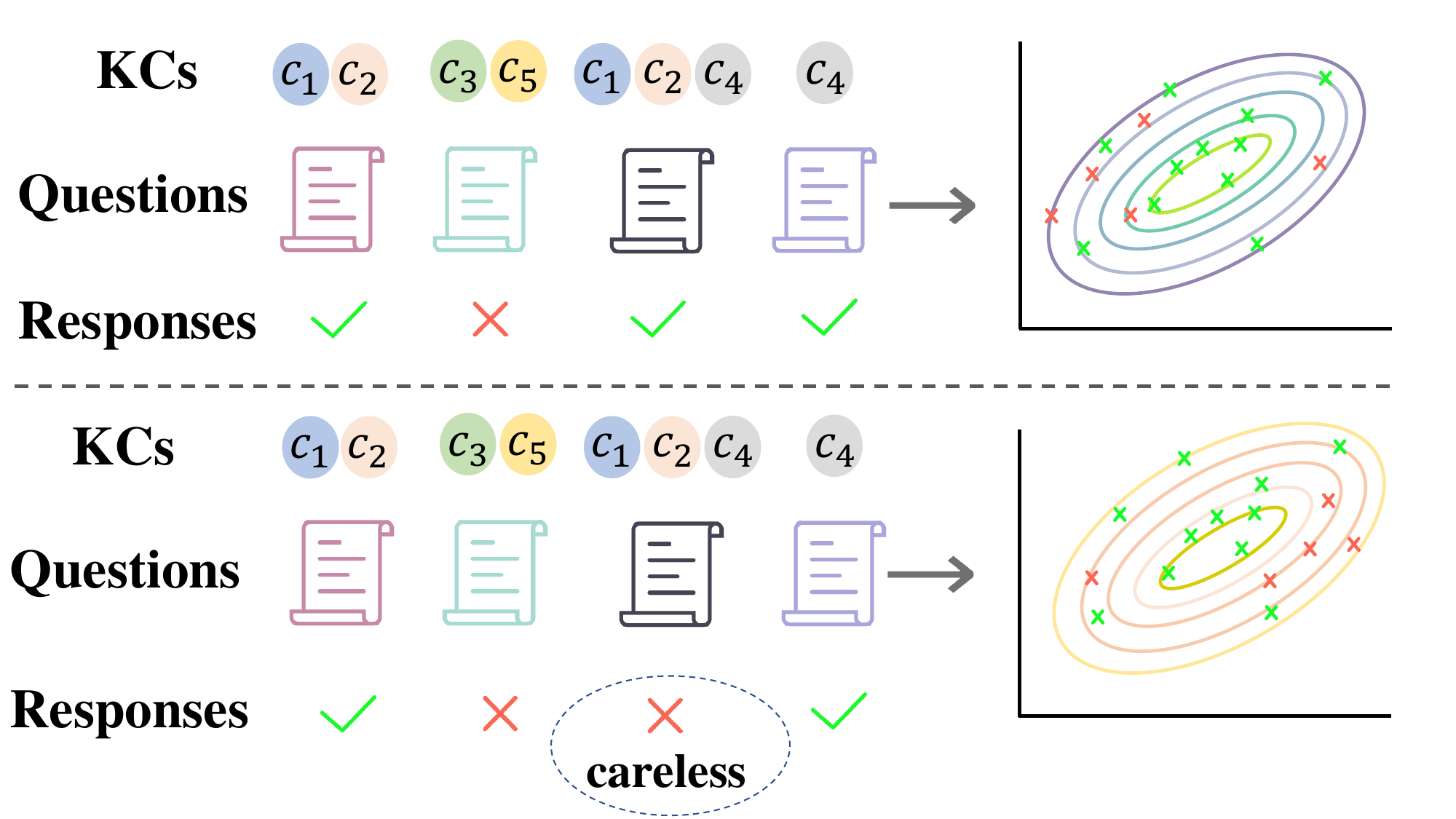}
		\caption{The impact of aleatory uncertainty on knowledge state under stochastic modeling. KCs are the knowledge concepts involved in the given questions.}
        \label{aleatory}
\end{figure}

Note that the uncertainties captured before contain both the epistemic and aleatory ones. To prevent the model from being misled by aleatory uncertainty (as illustrated in Figure \ref{aleatory}) and enhance its robustness in predicting knowledge mastery, we developed the aleatory uncertainty-aware contrastive learning mechanism. By constructing negative samples that reflect aleatory-uncertainty examples, we strengthen the model's robustness against biased distributions. As a result, the model becomes less susceptible to anomalous behaviors and is better equipped to accurately capture students' knowledge mastery.

To the best of our knowledge, UKT is the first to explore uncertainty within the realm of Knowledge Tracing. Our key contributions are summarized as follows: 
\begin{itemize} 
    \item We propose to model students' base knowledge levels and uncertainties through stochastic interaction learning, where each student is represented as a Gaussian distribution with the mean and covariance embeddings representing the corresponding mastery degree and uncertainty.
    \item Through the construction of negative samples representing both careless and lucky guesses in contrastive learning, we offer a solution to retain epistemic uncertainty while enhancing model robustness to aleatory uncertainty.
    \item We conduct extensive experiments on six datasets. The results indicate that the proposed method outperforms existing baselines in performance. Additionally, through additional experiments and analysis, we demonstrate that our method can effectively model epistemic uncertainty and exhibits strong robustness to aleatory uncertainty.
\end{itemize}

\section{Related Work}

Knowledge tracing aims to dynamically assess a learner's mastery of knowledge, thereby customizing subsequent learning paths.
The assessment is achieved by recording whether a student answers correctly to the given question. As a question commonly comprises specific knowledge points, the student's answer to the question would reflect her/his understanding level of involved knowledge concepts. The role of a knowledge tracing model is to predict whether a student can answer correctly to the given question based on the learning of historical question-answering records, and a better KT model should exhibit stronger prediction capability. Generally, knowledge tracing models can be broadly divided into five categories: 

\subsubsection{Deep Sequential Models} These deep sequential models for knowledge tracing use auto-regressive architectures to capture students' sequential interactions. For example, \cite{piech2015deep} introduces the Deep Knowledge Tracing (DKT) model, utilizing an LSTM layer to estimate knowledge mastery. \cite{lee2019knowledge} enhances DKT by incorporating a skill encoder that combines student learning activities with knowledge component (KC) representations.

\subsubsection{Memory-Augmented Models} These models leverage memory networks to capture latent relationships between KCs and students' knowledge states. \cite{zhang2017dynamic} utilized a static key memory matrix to store KC relationships and a dynamic value memory matrix to predict students' knowledge mastery levels.

\subsubsection{Adversarial Learning-Based Models} These models employ adversarial techniques to generate perturbations that improve the model's generalization capability. For instance, \cite{guo2021enhancing} proposes an attentive LSTM KT model jointly trained with both original and adversarial examples.

\subsubsection{Graph-Based Models} These methods use graph neural networks to model intrinsic relationships among questions, KCs, and interactions. \cite{liu2020improving} introduces a question-KC bipartite graph to explicitly capture the inner relations at the question, KC levels, and question difficulties. \cite{yang2021gikt} employs a graph convolutional network to represent the correlations between questions and KCs.

\subsubsection{Attention-Based Models} These models utilize the attention mechanism to capture dependencies between interactions. For example, SAKT~\cite{pandey2019self} applies a self-attention network to capture the relevance between KCs and students’ historical interactions. SAINT~\cite{choi2020towards} introduces an encoder-decoder structure to represent exercise and response embedding sequences, while AKT~\cite{ghosh2020context} introduces three self-attention modules to explicitly model students' forgetting behaviors using a monotonic attention mechanism. Furthermore, SimpleKT~\cite{liu2023simplekt} uses an ordinary dot-product attention to extract time-aware information embedded in student learning interactions.

However, existing research often overlooks the issue of uncertainty in knowledge tracing, which is a key challenge that needs to be addressed in the field.

\section{Preliminaries}

\subsection{Uncertainty Definition}
Uncertainty, as an inherent unpredictable component in model predictions or outcomes, plays a crucial role in supporting societal decision-making, quantitative research, and machine learning applications \cite{Helton2008, wimmer2023quantifying, gal2016uncertainty}. The prerequisite for delving into these uncertainties consist of establishing the foundation for building user trust in the system and enhancing the model reliability \cite{Sanchez2022}. To further refine this concept, uncertainty can be divided into two main categories: epistemic uncertainty and aleatory uncertainty \cite{gal2016uncertainty, hullermeier2021aleatoric}.

\subsubsection{Epistemic Uncertainty}  It arises from a lack of knowledge about the system or process and can be reduced by acquiring more information or improving the model, involving an incomplete understanding of certain parameters or system behaviors~\cite{Chen2020}. In knowledge tracing, epistemic uncertainty originates from the diversity in students' learning abilities and knowledge absorption levels and arises from varying depths of students' understanding of concepts so that is hard to capture. This uncertainty relates to each student's unique learning path, comprehension skills, and learning strategies. Even when faced with the same learning materials and environment, different students may exhibit significant differences in knowledge mastery and application, leading to variations in learning outcomes and complexity in assessing students' knowledge levels.

\subsubsection{Aleatory Uncertainty} This type of uncertainty is inherent in the data itself, stemming from the randomness in the data generation process. It cannot be reduced by collecting more data~\cite{hullermeier2021aleatoric}. Examples include measurement noise, inherent variability in the data, and randomness in the environment. In the field of knowledge tracing, this uncertainty might manifest as students making careless mistakes or guessing on questions.

By understanding and quantifying these two types of uncertainty, we can build models that are both accurate and trustworthy \cite{abdar2021review}. We have carefully designed our KT model to effectively tackle both epistemic and aleatory uncertainty.

\subsection{Problem Statement}
For each student \( S \), we assume that we have a series of \( T \) interactions arranged in chronological order, i.e., \( S = \{s_j\}_{j=1}^{T} \). Each interaction can be represented as a structured quadruple \( s = \langle q, \{c\}, r, s \rangle \), where \( q \) is the specific question, \( \{c\} \) is the set of knowledge components associated with the question, \( r \) is the binary student response (correct or incorrect), and \( s \) is the time point at which the student answered. Our goal is to construct a model based on this information to evaluate and predict the probability \( \hat{r}^* \) that a student will answer correctly to a given question \( q^* \).


\section{Methodology}

\begin{figure}
		\centering 
            \includegraphics[width=\linewidth,height=5cm,]{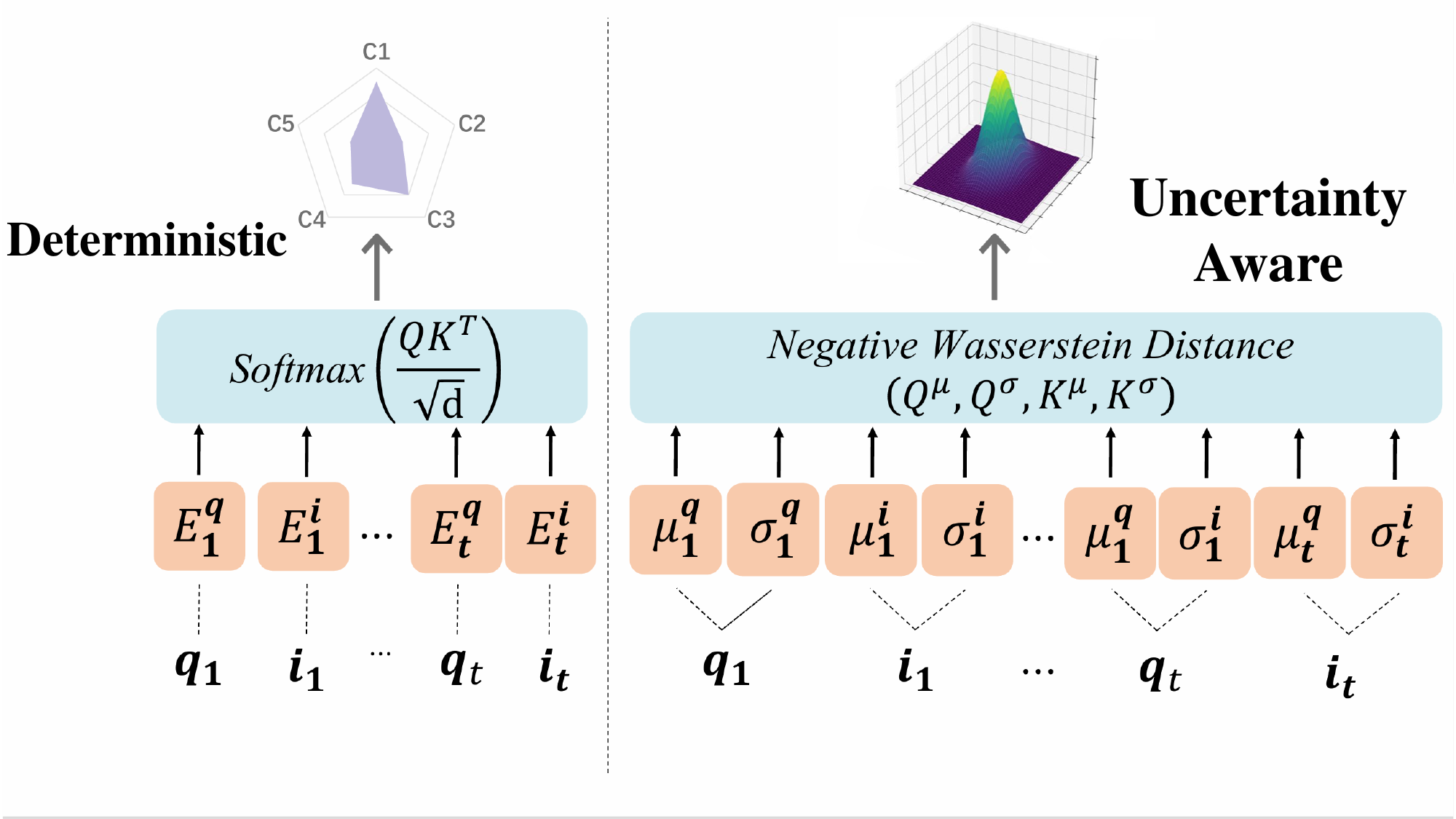}
		\caption{The differences between deterministic KT and UKT architectures.} \label{sample-figure}
\end{figure}
In this section, we describe how the UKT model processes student interaction data and handles uncertainty. The UKT model uses a stochastic embedding layer with mean and covariance parameters to represent student interactions as a Gaussian distribution. The mean indicates the student's baseline knowledge, while the covariance captures the uncertainty in the learning process, including both epistemic and aleatory uncertainties. A Wasserstein-based self-attention layer analyzes this Gaussian distribution to track changes in knowledge states. The student's knowledge state and associated uncertainty are then processed through a Feed-Forward Network (FFN) layer to assess their mastery and uncertainty regarding specific questions. To address aleatory uncertainty, we incorporate an aleatory uncertainty-aware contrastive learning layer for model learning to enhance prediction robustness. Finally, we integrate the adjusted uncertainty with the student's knowledge level to predict their performance on specific problems. The main difference between UKT and deterministic KT in model architectures is shown as figure \ref{sample-figure}.
\subsection{Stochastic Embedding Layers}
Understanding student interactions is crucial for tracking their learning progress. In education, practice questions often exceed knowledge components (KCs) significantly. For example, in the Algebra2005 dataset, questions outnumber KCs by over 1,500 times (see Section 4.1). 
To handle the inherent uncertainty, we build on the methods from \cite{ghosh2020context} and \cite{liu2022pykt}. We develop our UKT from \cite{ghosh2020context} and \cite{liu2022pykt}. Specifically, 
we first map students' responses to questions onto KCs, evaluating each KC individually to assess understanding. We then use mean embedding $E^\mu \in \mathbb{R}^{|V| \times d}$ to represent base interaction and covariance embedding $E^\sigma \in \mathbb{R}^{|V| \times d}$ for uncertainties, creating a multidimensional elliptical Gaussian distribution to student behavior.  We apply the same method to the knowledge components (KCs) as well to obtain $M^\mu \in \mathbb{R}^{|V| \times d}$,$M^\sigma \in \mathbb{R}^{|V| \times d}$. The representation process is formulated as follows:
\begin{align}
    z_{ck} &= \mathbf{W}_q \cdot e_{ck}, \quad r^{\sigma}_{qj} = \mathbf{W}^1_c \cdot e^{\sigma}_{qj}, \quad r^{\mu}_{qj} = \mathbf{W}^2_c \cdot e^{\mu}_{qj} \nonumber\\
        E^{\sigma}_t &= z_{ck} \oplus r^{\sigma}_{qj} \odot v_{ck}, \quad E^{\mu}_t = z_{ck} \oplus m^{\mu}_{qj} \odot v_{ck} \nonumber\\
        M^{\sigma}_t &= z_{ck} \oplus m^{\sigma}_{qj} \odot v_{ck}, \quad M^{\mu}_t = z_{ck} \oplus m^{\mu}_{qj} \odot v_{ck}
\end{align}
\noindent where \(z_{ck}\) denotes the latent representation of KC \(c_k\), \(m_{qj}\) represents the difficulty vector of the question \(q_j\) containing \(c_k\), and \(v_{ck}\) indicates the question-centric variation of \(q_j\) including \(c_k\). \(r_{qj}\) denotes the student's response representation to \(q_j\). \(e_{ck}\) and \(e_{qj}\) are one-hot vectors indicating the corresponding KC and correctness of the response, respectively. \(z_{ck}\), \(m_{qj}\), \(v_{ck}\), and \(r_{qj}\) are \(d\)-dimensional learnable vectors, and \(W_c \in \mathbb{R}^{d \times n}\) and \(W_q \in \mathbb{R}^{d \times 2}\) are learnable linear transformations. The operators \(\odot\) and \(\oplus\) denote element-wise product and addition, with \(n\) represents the total number of KCs.

\subsection{Wasserstein-Based Self-Attention}

To address the limitation of dot-product that can not measure the discrepancy between distributions \cite{kim2021lipschitz}, we introduce the Wasserstein distance \cite{clement2008elementary,fan2022sequential,fan2023mutual} to track the progression of a student's learning states across stochastic embeddings, in which we employ separate position embeddings for the mean and covariance, denoted as \(P^\mu \in \mathbb{R}^{n \times d}\) and \(P^\sigma \in \mathbb{R}^{n \times d}\), respectively:
\begin{align}
    &\hat{\mathbf{M}}^{\mu}_{t} = 
    \begin{bmatrix}
    \mathbf{M}^{\mu}_{t_1}+\mathbf{P}^{\mu}_{t_1}, & \mathbf{M}^{\mu}_{t_2}+\mathbf{P}^{\mu}_{t_2}, & \ldots, & \mathbf{M}^{\mu}_{t_n}+\mathbf{P}^{\mu}_{t_n}
    \end{bmatrix} \nonumber\\
    &\hat{\mathbf{M}}^{\sigma}_{t} =
    \begin{bmatrix}
    \mathbf{M}^{\sigma}_{t_1}+\mathbf{P}^{\sigma}_{t_1}, & \mathbf{M}^{\sigma}_{t_2}+\mathbf{P}^{\sigma}_{t_2}, & \ldots, & \mathbf{M}^{\sigma}_{t_n}+\mathbf{P}^{\sigma}_{t_n}
    \end{bmatrix}\nonumber\\
    &\hat{\mathbf{E}}^{\mu}_{t} = 
    \begin{bmatrix}
    \mathbf{E}^{\mu}_{t_1}+\mathbf{P}^{\mu}_{t_1}, & \mathbf{E}^{\mu}_{t_2}+\mathbf{P}^{\mu}_{t_2}, & \ldots, & \mathbf{E}^{\mu}_{t_n}+\mathbf{P}^{\mu}_{t_n}
    \end{bmatrix} \nonumber\\
    &\hat{\mathbf{E}}^{\sigma}_{t} =
    \begin{bmatrix}
    \mathbf{E}^{\sigma}_{t_1}+\mathbf{P}^{\sigma}_{t_1}, & \mathbf{E}^{\sigma}_{t_2}+\mathbf{P}^{\sigma}_{t_2}, & \ldots, & \mathbf{E}^{\sigma}_{t_n}+\mathbf{P}^{\sigma}_{t_n}
    \end{bmatrix}
\end{align} 
We derive the mean and covariance sequence embeddings for student interactions. To align with the interaction distribution, we also derive the mean and covariance embeddings of KC sequences.The Exponential Linear Unit (ELU) maps inputs into the interval $[-1, +\infty)$, ensuring the positive definite property of covariance.
\begin{align}
M^{\mu}_{t}      &= \hat{M}^{\mu}, 
&M^{\sigma}_{t}   &= \text{ELU} \left( \text{diag}(\hat{M}^{\sigma}) \right) + 1\nonumber\\
E^{\mu}_{t}      &= \hat{E}^{\mu},
&E^{\sigma}_{t}   &= \text{ELU} \left( \text{diag}(\hat{E}^{\sigma})\right) + 1
\end{align}
We apply the attention weight as the negative 2-Wasserstein distance $W_2(\cdot, \cdot)$ to knowledge state retrieval, which is computed as follows:
\begin{align}
&\mathbf{score}_{t+1} =  - \left( W_2(M_{t}, R_{t}) \right) =- \bigl( \left\| M^{\mu} - E^{\mu} \right\|^2_2 \nonumber\\
                       &+ \text{trace} \bigl( M^{\sigma} \!+ E^{\sigma} - 2 \bigl( (M^{\sigma})^\frac{1}{2} M^{\sigma} (E^{\sigma})^\frac{1}{2} \bigr)^\frac{1}{2} \bigr) \bigr)
\end{align}
and the retrieved knowledge state $\mathbf{h}_{t+1}$ at timestamp $t+1$ is computed as follows:
\begin{align}
&\mathbf{h}_{t+1}^{\mu},\mathbf{h}_{t+1}^{\sigma}  = \mbox{WassersteinSelfAttention}(  \notag\\ 
&Q^{\mu} = \mathbf{M^{\mu}}_{t+1},\quad  \quad  \quad \quad Q^{\sigma} = \mathbf{M^{\sigma}}_{t+1}, \nonumber\\
&K^{\mu} = \{\mathbf{M}^{\mu}_1, \cdots,\mathbf{M^{\mu}}_t\}, K^{\sigma} = \{\mathbf{M}^{\sigma}_1, \cdots,\mathbf{M^{\sigma}}_t\} , \nonumber\\
&V^{\mu} = \{\mathbf{E}^{\mu}_1, \cdots, \mathbf{E^{\mu}}_t\}, \quad V^{\sigma} = \{\mathbf{E}^{\sigma}_1, \cdots, \mathbf{E^{\sigma}}_t\}).
\end{align}
\subsection{Feed-Forward Network}
Then we use a two-layer fully connected network to refine the knowledge state:
\begin{equation}
\begin{aligned}
\textbf{h}^{\mu}_{t+1}\!= & \mathbf{W}_2 \cdot \mbox{ReLU} ( \mathbf{W}_1 \cdot [\mathbf{h}^{\mu}_{t+1}; \mathbf{M}^{\mu}_{t+1}] + \mathbf{b}_1 ) + \mathbf{b}_2   \\ 
\textbf{h}^{\sigma}_{t+1}\!= &\mbox{ELU}\bigl(\mathbf{W}_4\!\cdot \mbox{ReLU} ( \mathbf{W}_3\!\cdot\![\mathbf{h}^{\sigma}_{t+1}; \mathbf{M}^{\sigma}_{t+1}]\!+\! \mathbf{b}_3)\!+\!\mathbf{b}_4 \bigl)+1
\end{aligned}
\end{equation}
where $\mathbf{W}_1, \mathbf{W}_3 \in \mathbb{E}^{d \times 2d}$, $\mathbf{W}_2, \mathbf{W}_4 \in \mathbb{E}^{d \times d}$, and $\mathbf{b}_1, \mathbf{b}_2,\mathbf{b}_3,\mathbf{b}_4  \in \mathbb{E}^{d \times 1}$ are trainable parameters. 

The framework design of UKT is highly efficient. Compared to standard attention-based KT methods~\cite{liu2023simplekt}, UKT only adds a covariance matrix embedding, while Wasserstein attention has a computational cost similar to that of the standard self-attention. In our experiments, training and inference can be efficiently completed on a single GPU within an hour.

\subsection{	Aleatory Uncertainty-Aware Contrastive Learning}
To retain epistemic uncertainty and enhance the model's resilience to aleatory uncertainty, we propose aleatory uncertainty-aware contrastive learning, which constructs negative samples that represent both careless behavior and lucky guesses for contrastive learning. We denote the negative interaction sequence of student \( s_i \) as \( q_{s_i^{-}} \). Constructing such semantically negative samples is crucial for contrastive learning \cite{gao2021simcse, li2024multi}. Specifically, we modify a student's interaction sequence based on the outcome of the last answered question. If the last item is correct, we invert all previous correct responses and reserve the last correct one to construct the negative sequence, preventing model from being influenced by lucky guesses. Conversely, if the last interaction is incorrect, we invert previous incorrect responses with the last one unchanged as another kind of negative sample to prevent the model from overemphasizing careless mistakes. 
We adopt the categorical cross-entropy loss \cite{oord2018representation} to maximize mutual information (\( q^{s_i} \), \( q^{s_i^{-}} \)) between two perturbed views \cite{ozair2019wasserstein}, calculated as:

{\small
\begin{align}
&I_{W_2}(q^{s_i}, q^{s_i^{-}}) \geq -\mathcal{L}_{\text{CL}}(\textbf{h}^{s_i}, \textbf{h}^{s_i^{-}}) \nonumber \\
& = \log \frac{\exp\left(W_2(\textbf{h}^{s_i},\textbf{h}^{s_i^{-}})\right)}{\exp\left(W_2(\textbf{h}^{s_i},\textbf{h}^{s_i^{-}})\right)+\sum\limits_{j\in S_{\mathcal{B}}^-}\exp\left(-W_2(\textbf{h}^{s_i}, \textbf{h}^j)\right)} 
\end{align}
}where $\textbf{h}^{s_i} \sim \mathcal{N}\left(\textbf{h}^{s_i}_{\mu}, \textbf{h}^{s_i}_{\Sigma}\right)$ is the encoded stochastic output knowledge state with mean $\textbf{h}^{s_i}_{\mu}$ and covariance $\textbf{h}^{s_i}_{\Sigma}$. $I_{W_2}(q^{s_i}, q^{s_i^{-}})$ denotes the mutual information of ($q^{s_i}$ ,$q^{s_i^{-}}$). The 2-Wasserstein distance between the encoded distributions is computed as the sum of the squared $L_2$ distances between the mean embeddings and the squared root of the covariance matrices.

\subsection{Prediction Layer and Model Optimization}
The overall optimization function $\mathcal{L}_{overall}$ is defined as:
\begin{align}
& \eta_{t+1}  = \mathbf{w}^\top \cdot \mbox{ReLU} \bigl(h^{\mu}_{t+1}; h^{\sigma}_{t+1}\bigl) + b\\
& \mathcal{L}_{p}  = - \sum\nolimits_t \bigl( r_t \log \sigma(\eta_t) + (1-r_t) \log (1-\sigma(\eta_t)) \bigl)\\
& \mathcal{L}_{overall}  = \mathcal{L}_{p}  + \lambda * \mathcal{L}_{\text{CL}}
\end{align}
\noindent where $\mathbf{w}$ and $b$ are trainable parameters and $\mathbf{w} \in \mathbb{E}^{d \times 1}$, $b$ is the scalar, $\sigma(\cdot)$ is the Sigmoid function and $\lambda$ is the weight of contrastive learning loss.
\section{Experiments}
In this section, we perform extensive experiments to answer the following Research Questions (\textbf{RQ}s):
\begin{itemize}
    \item\textbf{RQ1}: How does UKT perform compared with baselines?

\item\textbf{RQ2}: How does UKT perform with different weights for aleatory uncertainty-aware contrastive learning?

\item\textbf{RQ3}: Can UKT capture epistemic uncertainties and reduce the effect of aleatory uncertainties for KT?

\item\textbf{RQ4}: Are the key components effective in UKT?
\end{itemize}
\begin{table*}[!bpht]
\small
\centering
\begin{tabular}{c|cccccc}
\toprule

\multirow{1}{*}{\textbf{Model}}    
                  & \textbf{AS2009}         & \textbf{AL2005}        & \textbf{BD2006}        & \textbf{NIPS34}         
                  & \textbf{AS2015}         & \textbf{POJ}           \\ \hline

\textbf{DKT}      & 0.8226±0.0011         & 0.8149±0.0011            & 0.8015±0.0008             & 0.7689±0.0002         
                  & 0.7271±0.0005         & 0.6089±0.0009    \\      
\textbf{SAKT}     & 0.7746±0.0017           & 0.8780±0.0063          & 0.7740±0.0008          & 0.7517±0.0005         
                  & 0.7114±0.0003           & 0.6095±0.0013          \\
\textbf{SAINT}    & 0.7458±0.0023           & 0.8775±0.0017          & 0.7781±0.0013          & 0.7873±0.0007        
                  & 0.7026±0.0011           & 0.5563±0.0012          \\

\textbf{ATKT}    & 0.7470±0.0008           & 0.7995±0.0023          & 0.7889±0.0008          & 0.7665±0.0001        
                  & 0.7245±0.0007           & 0.6075±0.0012          \\

\textbf{AKT}      & 0.8474±0.0017           & 0.9294±0.0019          & 0.8167±0.0007          &  0.7960±0.0003
                  & \textbf{0.7282±0.0004}  & 0.6218±0.0013          \\
                  
\textbf{simpleKT} & 0.8413±0.0018           & 0.9267±0.0003          & 0.8141±0.0006          &  0.7966±0.0000   
                  & 0.7237±0.0005           & 0.6194±0.0005          \\      \hline                 

\textbf{UKT}      & \textbf{0.8563±0.0014}  & \textbf{0.9320±0.0012}   & \textbf{0.8178±0.0009}          & \textbf{0.8035±0.0004}  
                  & 0.7267±0.0007           & \textbf{0.6301±0.0005} \\

\bottomrule
\end{tabular}
\caption{Overall AUC performance of \textsc{UKT} and all baselines. } 
\label{tab:overall_auc}

\end{table*}

\begin{table*}[!bpht]
\small

\centering
\begin{tabular}{c|cccccc}
\toprule
\multirow{1}{*}{\textbf{Models}} 
                  & \textbf{AS2009}        & \textbf{AL2005}        & \textbf{BD2006}        & \textbf{NIPS34}         
                  & \textbf{AS2015}        & \textbf{POJ}              
                  \\ \hline
\textbf{DKT}      &  0.7657±0.0011         & 0.8149±0.0011         & 0.8015±0.0008         & 0.7689±0.0002          
                  & 0.7271±0.0005          & 0.6089±0.0009         \\
\textbf{SAKT}     & 0.7063±0.0018          & 0.7954±0.0020          & 0.8461±0.0005          & 0.6879±0.0004          
                  & 0.7474±0.0002          & 0.6407±0.0035         
                  \\
\textbf{SAINT}    & 0.6936±0.0034          & 0.7791±0.0016          & 0.8411±0.0065          & 0.7180±0.0006         
                  & 0.7438±0.0010          & 0.6476±0.0003         
                  \\

\textbf{ATKT}    & 0.7208±0.0009          & 0.7998±0.0019          & 0.8511±0.0004          & 0.6332±0.0023        
                  & 0.7494±0.0002        & 0.6075±0.0012          \\
\textbf{AKT}      & 0.7772±0.0021          & 0.8747±0.0011          & 0.8516±0.0005          & 0.7323±0.0005       
                  & \textbf{0.7521±0.0005} & 0.6449±0.0010          \\  
\textbf{simpleKT} & 0.7748±0.0012          & 0.8510±0.0005          & 0.8510±0.0003          &0.7328±0.0001 
                  & 0.7506±0.0004          & 0.6498±0.0008          \\ \hline 
                  
\textbf{UKT}      & \textbf{0.7814±0.0017} & \textbf{0.8781±0.0005} & \textbf{0.8531±0.0006}  & 0.7316±0.0004        
                  & 0.7497±0.0002          & \textbf{0.6548±0.0008}                 \\
\bottomrule
\end{tabular}
\caption{Overall Accuracy performance of \textsc{UKT} and all baselines. } 
\label{tab:overall_acc}

\end{table*}

\subsection{Dataset}
In our experiments, we conduct experiments on six benchmark datasets to evaluate the performance of each model:

\begin{itemize}
    \item \textbf{ASSISTments2009 (AS2009)}\footnote{ \url{https://sites.google.com/site/assistmentsdata/home/2009-2010-assistment-data/skill-builder-data-2009-2010}. }: This dataset focuses on math exercises, collected from the free online tutoring platform ASSISTments during the 2009-2010 school year. It has been widely used as a standard benchmark for KT methods over the past decade. The dataset includes 3,374,115 interactions, 4,661 sequences, 17,737 questions, and 123 knowledge components (KCs), with each question having an average of 1.1968 KCs.

    \item  \textbf{Algebra2005 (AL2005)} \footnote{\label{ft:al_bd}\url{https://pslcdatashop.web.cmu.edu/KDDCup/}.}: This dataset originates from the KDD Cup 2010 EDM Challenge and includes detailed step-level student responses to mathematical problems, where a question is constructed by concatenating the problem name and step name. This dataset consists of 884,098 interactions, 4,712 sequences, 173,113 questions, and 112 KCs, with an average of 1.3521 KCs per question.
    
    \item  \textbf{Bridge2006 (BD2006)}:
    It is derived from the KDD Cup 2010 EDM Challenge as well, this dataset follows a similar unique question construction process as used in Algebra2005. The dataset contains 1,824,310 interactions, 9,680 sequences, 129,263 questions, and 493 KCs, with an average of 1.0136 KCs per question.
    
    \item \textbf{NIPS34}\footnote{\url{https://eedi.com/projects/neurips-education-challenge}.}: Provided by the NeurIPS 2020 Education Challenge, this dataset contains students' answers to mathematics questions from Eedi. The dataset used from Task 3 and Task 4 includes 1,399,470 interactions, 9,401 sequences, 948 questions, and 57 KCs, with each question having an average of 1.0137 KCs.

    \item \textbf{ASSISTments2015 (AS2015)}\footnote{ \url{https://sites.google.com/site/assistmentsdata/datasets/2015-assistments-skill-builder-data}.}: Similar to ASSISTments2009, this dataset is collected from the ASSISTments platform in 2015 and includes the largest number of students among the other ASSISTments datasets. After pre-processing, it includes 682,789 interactions, 19,292 sequences, and 100 KCs.
    
    \item \textbf{POJ}\footnote{ \url{https://drive.google.com/drive/folders/1LRljqWfODwTYRMPw6wEJ_mMt1KZ4xBDk}. }: Collected from the Peking University coding practice online platform, this dataset includes 987,593 interactions, 20,114 sequences, and 2,748 questions.
        
\end{itemize}

Following the data pre-processing steps suggested by \citep{liu2022pykt}, we remove student sequences with fewer than three attempts and set the maximum length of student interaction history to 200 to ensure computational efficiency.

\subsection{Baselines}
We evaluate the performance of the proposed UKT by comparing it with the following baselines:

\begin{itemize}
    \item \noindent \textbf{DKT}  \citep{piech2015deep}: DKT directly uses RNNs to model students’ learning processes.
     \item \noindent \textbf{SAKT} \citep{pandey2019self}: It employs self-attention to identify the relevance between the
    interactions and KCs.
    
     \item \noindent \textbf{SAINT} \citep{choi2020towards}: It is a Transformer-based model for KT that encodes exercise and
    responses in the encoder and decoder respectively.
    
    \item \noindent \textbf{ATKT} \citep{guo2021enhancing}: This approach utilizes adversarial perturbations to improve the generalization ability of the attention-LSTM-based knowledge tracing model.
    
     \item \noindent \textbf{AKT} \citep{ghosh2020context}: This approach models forgetting behaviors during relevance computation
    between historical interactions and target questions.

    
     \item \noindent \textbf{SimpleKT} \citep{liu2023simplekt}: It utilizes a simple and effective attention mechanism to capture the contextual information embedded in students' learning interactions.
\end{itemize}

\subsection{Implementation Details and Evaluation Metrics}

Following previous research~\cite{liu2022pykt}, we reserve 20\% of the students' sequences for evaluation, while the remaining 80\% undergo standard 5-fold cross-validation.

For model training,
we use the Adam optimizer~\cite{kingma2014adam}, capping the training at a maximum of 200 epochs with early stopping to speed up the process. The embedding dimension, hidden state dimension, and the two dimensions of the prediction layers are set within [64, 128, 256, 512]. Learning rates are selected from [1e-3, 1e-4, 1e-5], dropout rates from [0.05, 0.1, 0.3, 0.5], and contrastive learning rates from [0.01, 0.02, 0.05, 0.07, 0.1, 0.5, 1]. The number of blocks and attention heads are set within [1, 2, 4] and [4, 8], respectively. Our model is implemented with PyTorch and trained on a single A100 GPU.

In line with previous research, we evaluate model performance with AUC and accuracy as metric.

\subsection{Overall Performances (RQ1)}
The model performance in terms of the average AUC and accuracy scores is reported in Table \ref{tab:overall_auc} and Table \ref{tab:overall_acc} respectively. We can draw the following conclusions:
\begin{itemize}
    \item \textsc{UKT} consistently outperforms the other baselines in AUC scores across all datasets, highlighting its superiority as an uncertainty-aware KT model.
    \item \textsc{UKT} only slightly underperforms baselines on the ASSIST2015 dataset. This can be attributed to the fact that the ASSIST2015 dataset is the largest and primarily consists of data from different students, allowing AKT's two-layer attention structure to simultaneously ignore individual differences in epistemic uncertainty and aleatory uncertainty, leveraging the large dataset for predictions.
    \item  \textsc{UKT} demonstrates significant advantages over other baselines across all datasets. This is particularly evident in terms of AUC and Accuracy, where the improvements are substantial. This further validates that by focusing on the uncertainty in students' learning states, \textsc{UKT} can more effectively capture and predict learning behaviors, thereby excelling across various datasets.
\end{itemize}

\begin{figure}[b] 
		\centering 
            \includegraphics[width=\linewidth]{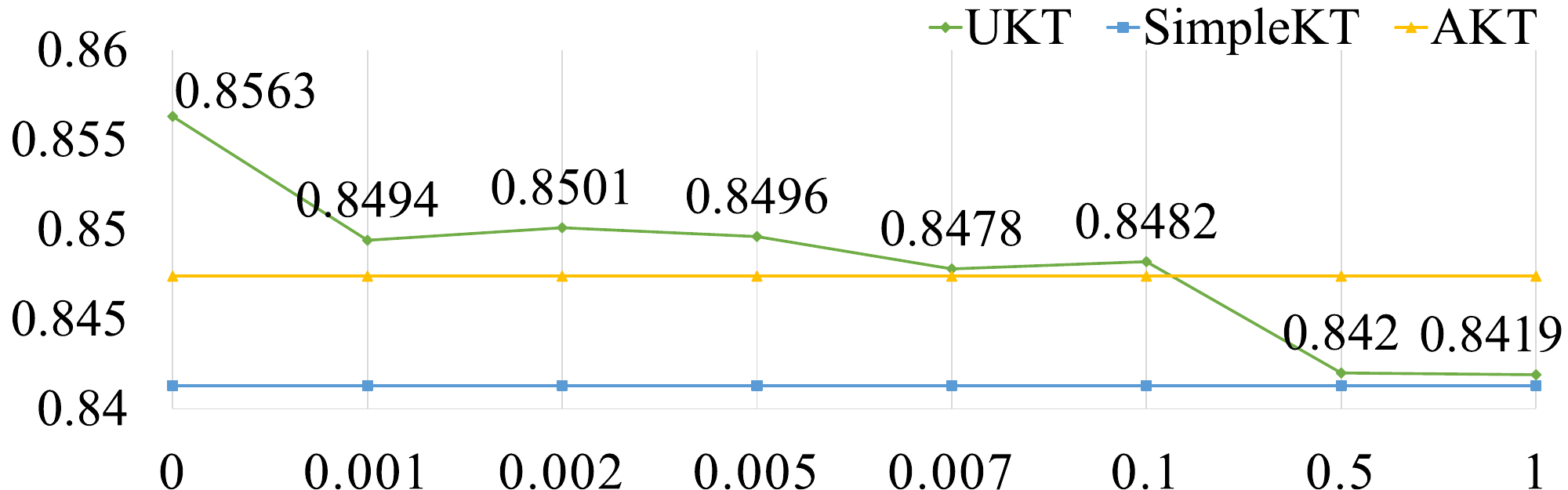}
		\caption{AUC of UKT with varying $\lambda$ values.} \label{lambda}
\end{figure}

\subsection{Contrastive Learning Weight Analysis (RQ2)}
To explore the impact of the degree of emphasis on aleatory uncertainty on the overall model performance, we adjusted the weight of contrastive learning to observe the model's final prediction performance. 
Figure \ref{lambda} shows the performance on ASSIST2009 and Algebra2005 with varying $\lambda$ values, from which we draw the following conclusions:
\begin{itemize} 
    \item Aleatory uncertainty is a random factor in students' learning assessment. 
    By mitigating the effect of this uncertainty, we can indeed enhance the performance.
    \item When too much emphasis is placed on aleatory uncertainty, the model's performance actually declines. 
    The reason is that aleatory uncertainty is more sparse and random compared to epistemic uncertainty. Excessive weight $\lambda$ would bring too much attention to aleatory uncertainty during training, diverting the model from the primary goal of predicting student knowledge levels.
\end{itemize}

\subsection{Uncertainty Analysis (RQ3)}

\begin{figure}[t] 
		\centering 
            \includegraphics[width=\linewidth,height=5cm]{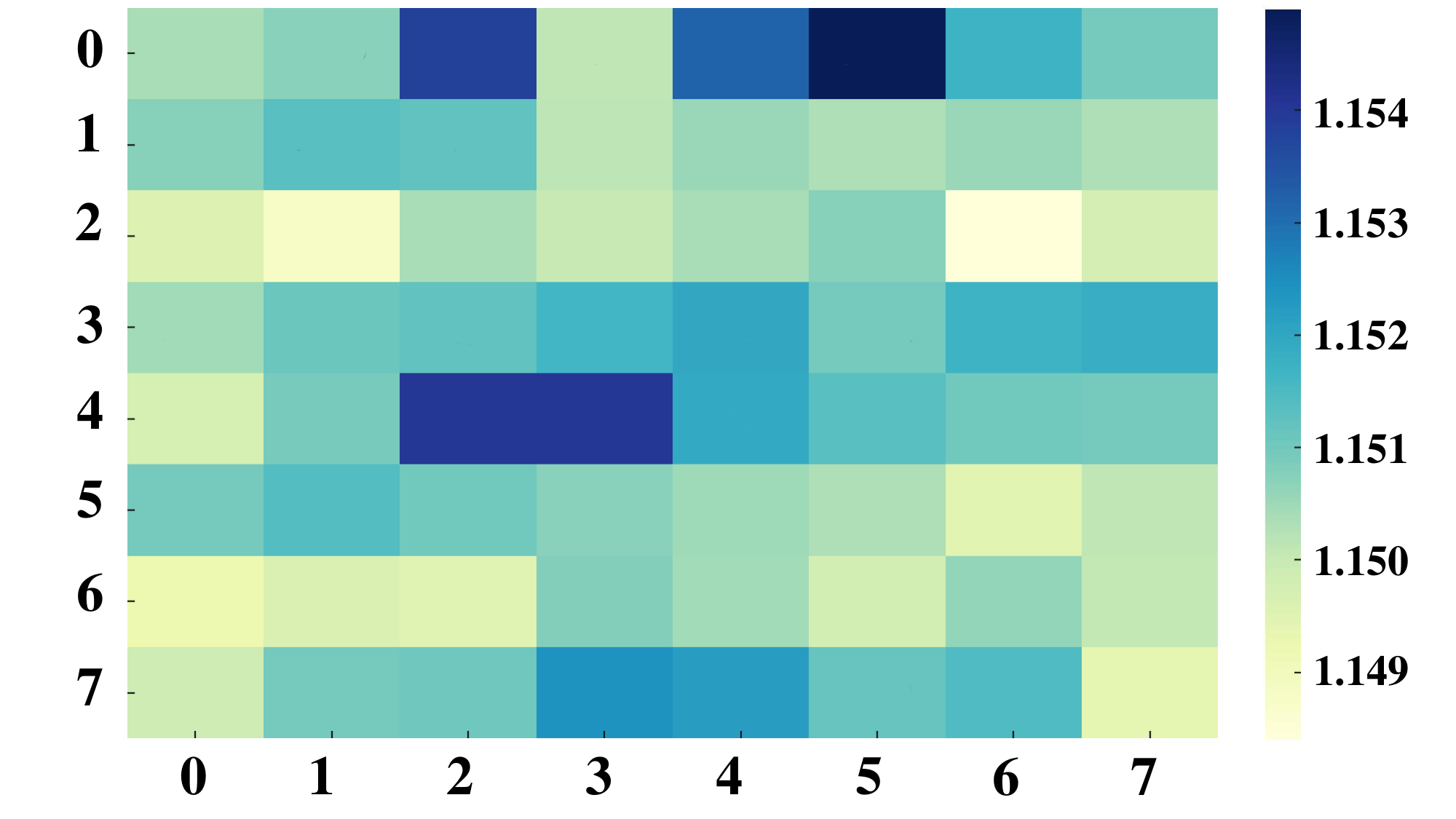}
		\caption{The heatmap of the mean of the covariance matrix of sequences in a batch from the Algebra2005 dataset.} \label{low_diff}
\end{figure}

\begin{figure}[t]
		\centering 
            \includegraphics[width=\linewidth,height=5cm]{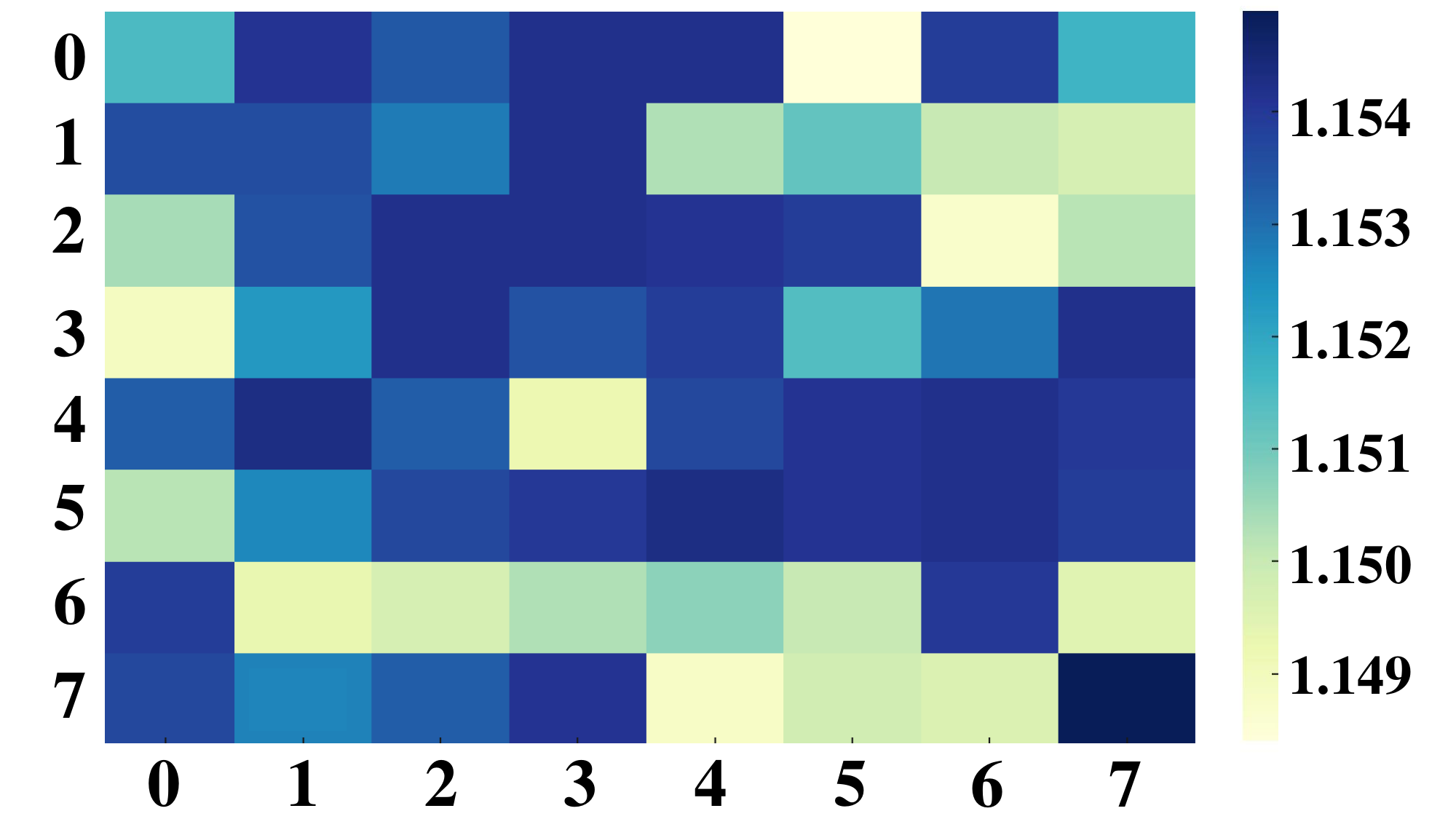}
		\caption{The heatmap of the mean of the covariance matrix of sequences in a batch from the Assist2009 dataset.} \label{high_diff}
\end{figure}

In this section, we verify that UKT is capable of capturing epistemic uncertainty while handling aleatory uncertainty through visualization.

We speculate that the epistemic uncertainty exhibited by the same student in a problem-solving scenario has a certain degree of continuity. Given that covariance embedding can quantify this uncertainty, we employ the following approach: we extract covariance embeddings processed through a Feed-Forward Network from a series of data and calculated the average covariance matrix for each sample to reflect the level of epistemic uncertainty across the entire sequence. We expect that the interaction sequences from the same student would show consistent average covariance values, which will be reflected as similar color blocks in the heatmap black (See Figures \ref{low_diff} and \ref{high_diff}).

Figure \ref{low_diff} is derived from the Algebra 2005 dataset, which records students' problem-solving process on math problems. The heatmap shows that the color gradient between adjacent data points is relatively smooth, indicating that the changes in covariance are minimal. Such observation confirms the relative stability of students' cognitive activity.
In contrast, Figure~\ref{high_diff} is from the assist2009 dataset, where the sequences primarily come from students with diverse backgrounds. The results show that the color blocks on the heatmap display more significant independence and dispersion, indicating clear differences in epistemic uncertainty among different students. This comparative analysis highlights the individuality and diversity of cognitive strategies each student possesses when solving problems, supporting our theoretical hypothesis.

\begin{table}[t]
\centering
\small
\begin{tabular}{c|cc|c}
\toprule
\textbf{Models} & \textbf{w/o AU} & \textbf{with AU} & \textbf{Performance} \\ 
\midrule
\textbf{simpleKT} & 0.8413 & 0.8317 & -1.14\% \\  
\textbf{AKT}      & 0.8474 & 0.8361 & -1.33\% \\                    
\textbf{UKT}      & 0.8501 & 0.8430 & -0.83\% \\

\bottomrule
\end{tabular}
\caption{Performance comparison of models with and without AU (Aleatory Uncertainty) on AS2009 dataset.}
\label{woau}
\end{table}

Furthermore, to demonstrate that UKT is more capable of handling uncertain environments compared to similar models, we investigate whether the aleatory uncertainty would bring difference to the performance of UKT and the other two baselines SimpleKT and AKT. As shown in Table \ref{woau}, UKT is less affected by uncertain data, demonstrating its superior ability to handle uncertain environments.


\subsection{Ablation Study (RQ4)}

As shown by the ablation study reported in Table~\ref{tab:ablation}, removing each component in UKT leads to inferior performance, indicating the effectiveness of  each component in uncertainty modeling. Notably, even without CL, UKT still outperforms other models, demonstrating the advantages of uncertainty modeling for KT.

\begin{table}[t]
\small
\setlength\tabcolsep{4pt} 
\centering
\begin{tabular}{l|ccc}
\toprule
\textbf{Model} & \textbf{AS2009}  & \textbf{Algebra2005}                \\  \hline         
\textbf{UKT}      & 0.8563±0.0018   &0.9320±0.0012          \\     
\textbf{UKT w/o CL} & 0.8507±0.0020     &0.9258±0.0018         \\    
\textbf{UKT w/o W.dist} & 0.8450±0.0024    &0.9208±0.0019           \\   
\textbf{UKT w/o Stocemb} & 0.8445±0.0028   &0.9217±0.0022            \\   
\bottomrule
\end{tabular} 
\caption{The ablation study of contrastive learning.} 
\vspace{-0.3cm}
\label{tab:ablation}
\end{table}

\section{Conclusion}
We propose a novel approach to model students' knowledge levels and uncertainties using stochastic interaction learning, where students are represented as Gaussian distributions with mean and covariance embeddings. By constructing negative samples that account for both careless and lucky guesses in contrastive learning, we enhance the model's robustness to aleatory uncertainty. Extensive experiments on six datasets show that our method outperforms existing baselines and effectively models epistemic uncertainty while demonstrating strong resilience to aleatory uncertainty. In the future, we plan to expand our research in the multimodal field~\cite{ni2023content, fu2024iisan} to handle more types of uncertainties.

\bibliography{aaai25}



\end{document}